\documentclass[conference]{IEEEtran}
\IEEEoverridecommandlockouts

\usepackage{cite}
\usepackage{amsmath,amssymb,amsfonts}
\usepackage{algorithmic}
\usepackage{graphicx}
\usepackage{textcomp}
\usepackage{xcolor}
\usepackage{multirow}
\def\BibTeX{{\rm B\kern-.05em{\sc i\kern-.025em b}\kern-.08em
    T\kern-.1667em\lower.7ex\hbox{E}\kern-.125emX}}
\begin{document}

\title{Boundary-Driven Table-Filling with Cross-Granularity Contrastive Learning for Aspect Sentiment Triplet Extraction
}

\author{
\IEEEauthorblockN{Qingling Li\textsuperscript{\rm 1}, Wushao Wen\textsuperscript{\rm 1,$\star$}, Jinghui Qin\textsuperscript{\rm 2,3,$\star$}\thanks{\textsuperscript{$\star$} Wushao Wen and Jinghui Qin are corresponding authors.}\thanks{This work was supported in part by the National Natural Science Foundation of China (NSFC) under Grant No. 62206314 and Grant No. U1711264, GuangDong Basic and Applied Basic Research Foundation under Grant No. 2022A1515011835, Science and Technology Projects in Guangzhou under Grant No. 2024A04J4388.}}
    \IEEEauthorblockA{
    \textsuperscript{\rm 1}Sun Yat-Sen University \quad 
    \textsuperscript{\rm 2}Guangdong University of Technology\\
    \textsuperscript{\rm 3}Guangdong Shuye Intelligent Technology Co., Ltd.\\
    liqling23@mail2.sysu.edu.cn, wenwsh@mail.sysu.edu.cn, scape1989@gmail.com
    }
}


\maketitle

\begin{abstract}
The Aspect Sentiment Triplet Extraction (ASTE) task aims to extract aspect terms, opinion terms, and their corresponding sentiment polarity from a given sentence. It remains one of the most prominent subtasks in fine-grained sentiment analysis. Most existing approaches frame triplet extraction as a 2D table-filling process in an end-to-end manner, focusing primarily on word-level interactions while often overlooking sentence-level representations. This limitation hampers the model's ability to capture global contextual information, particularly when dealing with multi-word aspect and opinion terms in complex sentences. To address these issues, we propose boundary-driven table-filling with cross-granularity contrastive learning (BTF-CCL) to enhance the semantic consistency between sentence-level representations and word-level representations. By constructing positive and negative sample pairs, the model is forced to learn the associations at both the sentence level and the word level. Additionally, a multi-scale, multi-granularity convolutional method is proposed to capture rich semantic information better. Our approach can capture sentence-level contextual information more effectively while maintaining sensitivity to local details. Experimental results show that the proposed method achieves state-of-the-art performance on public benchmarks according to the F1 score.

\end{abstract}


\section{Introduction}
\begin{figure}[t] 
	\centering
	\begin{minipage}[t]{0.5\textwidth}
		\centering
		\includegraphics[height=3cm]{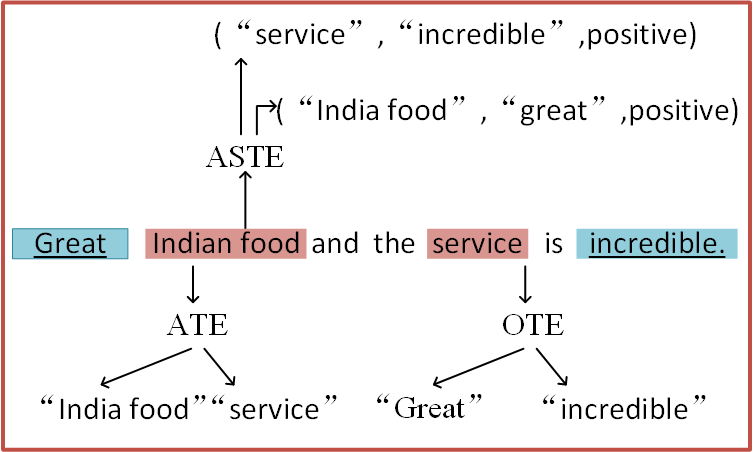}
	\end{minipage}
	\caption{An example of ABSA including ATE, OTE, and ASTE. The orange words represent aspects, and the blue ones represent opinions.}
        \label{fig1}
\end{figure}

Aspect-Based Sentiment Analysis (ABSA) is a fine-grained task that focuses on identifying the sentiments expressed toward specific aspect terms \cite{pontiki-etal-2014-semeval} \cite{pontiki-etal-2015-semeval} \cite{pontiki2016semeval}. Typically, to determine the sentiment corresponding to an aspect term, it is necessary to identify the sentiment-laden words associated with it in the sentence, referred to as opinion terms. Therefore, to accomplish the ABSA task, some subtasks have been introduced, including Aspect Term Extraction (ATE) \cite{manek2017aspect, li2018aspect, ma2019exploring, akhtar2020multi}, Opinion Term Extraction (OTE) \cite{wang2019transferable, wu2020deep, dai2022reasoning, feng2021target}, and more recently developed Aspect Sentiment Triplet Extraction (ASTE) \cite{peng2020knowing, xu2020position, zhang2020multi, xu2021learning}. Among these subtasks, ATE focuses on identifying and extracting aspect terms from a sentence, while OTE handles the extraction of opinion terms. ASTE combines both by identifying aspect terms, opinion terms, and their corresponding sentiment polarity expressed. Fig. \ref{fig1} illustrates examples of the different subtasks. 

The ASTE task was first proposed by Peng et al. \cite{peng2020knowing} which uses a two-stage pipeline approach. In the first stage, potential aspect and opinion terms are extracted as a labeling problem, while in the second stage, the aspect and opinion terms are paired, and a classifier is used to determine their corresponding sentiments.
Building on Peng et al.'s pipeline, many studies have investigated the interaction among subtasks to design the end-to-end triplet extraction models \cite{xu2020position, xu2021learning, wu2020grid, li2024improving, jing2021seeking, zhang2022boundary, sun2024rethinking}. Notably, in prior works such as \cite{wu2020grid}, \cite{jing2021seeking}, \cite{zhang2022boundary}, and \cite{sun2024rethinking}, the triplet extraction process is handled by using the table-filling method. In this approach, aspect and opinion terms are extracted from the diagonal elements of the table, while sentiment is represented as relational tags in the non-diagonal elements. The table-filling method enables comprehensive word-pair calculations and, being end-to-end, avoids the error propagation often seen in pipeline approaches. Additionally, this method simplifies the identification of relationships between words, making the process more efficient.

However, the table-filling method, which represents each triplet as a relation region in the 2D table and transforms the ASTE task into
detection and classification of relation regions, mainly centers on word-level interactions, emphasizing relationships between individual words. These methods excel at capturing fine-grained representations between individual words, but they often neglect to consider and incorporate sentence-level representations. This oversight limits the model's capacity to grasp global contextual information, which is crucial for understanding the full meaning of the entire sentence. As a result, when handling more complex sentences containing multi-word aspect and opinion terms, these models struggle to accurately capture the relationships between the terms and the overall sentiment, leading to potential performance drawbacks.

In this paper, based on the boundary-driven table-filling framework for aspect sentiment triplet extraction, we propose a cross-granularity contrastive learning (CCL) mechanism, which is designed to enhance the semantic consistency between the sentence-level representation and the word-level representations. We named our method as BTF-CCL. By constructing positive and negative sample pairs, the BTF-CCL is compelled to learn the intricate relationships and associations between the sentence-level representation and the word-level representations, improving its ability to capture both fine-grained and broad contextual information. This process ensures the model can better align local word-level details with the overall sentence-level meaning, leading to more accurate triplet extraction. In addition to cross-granularity contrastive learning, a multi-scale, multi-granularity convolutional method (MMCNN) is also introduced to enable the model to capture rich semantic information across different levels of granularity. Experimental results show that our method achieves better performance over existing state-of-the-art approaches.

\section{Proposed Method}
\subsection{Task Definition}
\begin{figure}[t]
	\centering
	\begin{minipage}[t]{0.5\textwidth}
		\centering
		\includegraphics[height=4cm]{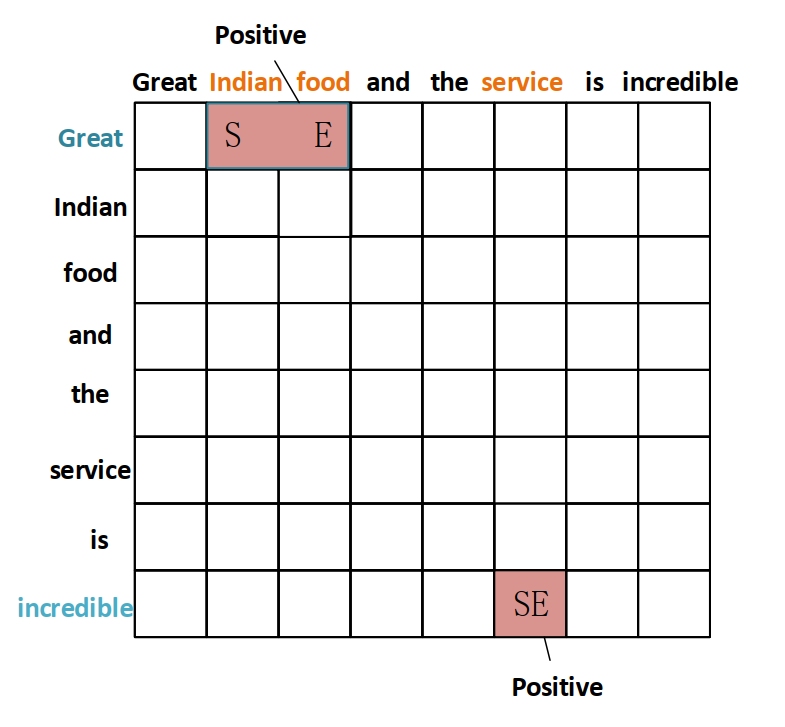}
	\end{minipage}
	\caption{An example of the aspect and opinion terms marked regions in the 2D table.}
        \label{fig2}
\end{figure}

$X=\{x_1,x_2,...,x_n\}$ represents a sentence consisting of $n$ words. The ASTE task aims to extract triplets represented as (aspect, opinion, sentiment), where sentiment can be categorized as Positive, Negative, or Neutral. As illustrated in Fig. \ref{fig2}, a triplet is depicted as a region within a 2D table. The boundaries of this region specify the position of the aspect term and the opinion term, while the type indicates the sentiment. The regions are marked by boundary tags: ‘S' for the upper left and ‘E’ for the lower right corner.

\subsection{Model Architecture }
\begin{figure*}[t]
	\centering
	\begin{minipage}[t]{0.82\textwidth}
		\centering
		\includegraphics[height=5cm]{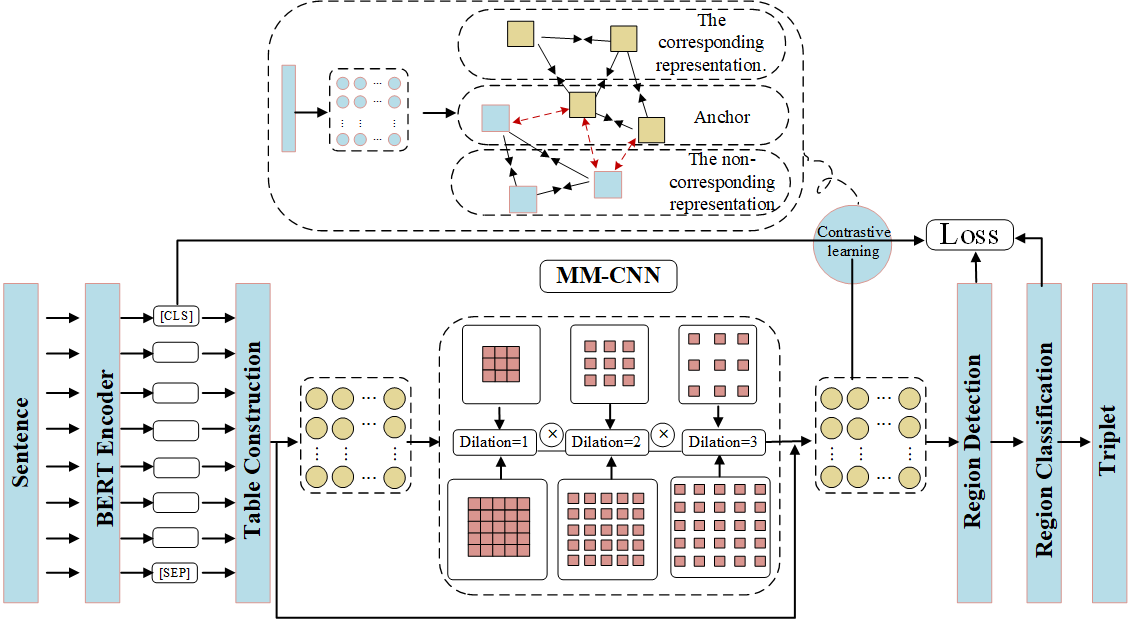}
	\end{minipage}
	\caption{The overview of BTF-CCL. The sentence is encoded by BERT Encoder, enriched with word-level representations via MMCNN, and then contrastive learning between sentence-level and word-level representations is applied. Finally, triplets are extracted via region detection and classification.}
        \label{fig3}
\end{figure*}

The architecture of our model BTF-CCL is shown in Fig. \ref{fig3}. We use BERT to encode input sentences and generate a 2D table representing word relationships. This table is further processed by a multi-scale, multi-granularity CNN (MMCNN) to capture local semantic information better. Cross-granularity contrastive learning is then applied between the word-level representations from the MMCNN and sentence-level representations, and all candidate regions are detected and classified for sentiment polarity. 

\subsubsection{Representation Leanrning}
For an input sentence $X=\{x_1, x_2, \dots, x_n\}$, we utilize the pre-trained BERT language model \cite{devlin2018bert} to generate contextual embeddings. By extracting the hidden states from the final layer of the encoder, we obtain the full sentence representation $H=\{h_1, h_2, \dots, h_n\}$.
 The relationship representation between two words $(h_i, h_j)\in R^d$ at position $(index_{h_i}, index_{h_j})$ in the table is generated through a nonlinear transformation, as shown in Equation \ref{equ1}, where $d$ is the hidden size of the BERT output.

 \begin{equation}
    r^{(0)}_{ij} = f(\text{Linear}([h_i; h_j; c_{ij}; t_{ij}]))
    \label{equ1}
\end{equation}
where $f(\cdot)$ is the activation function, and $[;]$represents the concatenation operation, $c_{ij}$ is the context representation between the word pair, obtained via max-pooling, and $t_{ij}$  represents the word-to-word interactions across different vector spaces, captured through a tensor-based operation.

\subsubsection{MMCNN}
The relation representations between word pairs come from a 2D matrix with dependencies, such as boundary tags `S' and `E'  and shared sentiment within the same region. This architecture applies a ResNet-style CNN with different convolution kernel sizes ($3\times3$, $5\times5$) and dilations (1, 2, 3) to table representations, referred to as multi-scale, multi-granularity convolutions (MMCNN), enabling the capture of richer local information and long-range dependencies.
\begin{equation}
    T' =  \text{ReLU}((\text{Conv}_{1 \times 1}(T^{(l-1)}))) 
    \label{equ2}
\end{equation}
\begin{equation}
    T'_{3x3, d1, d2, d3} = \text{ReLU}(\text{Conv}_{3 \times 3, d1, d2, d3}(T'))) 
    \label{equ3}
\end{equation}
\begin{equation}
    T'' =  T'_{3x3} + T'_{5x5}
    \label{equ6}
\end{equation}
\begin{equation}
    T''' = \text{ReLU}((\text{Conv}_{1 \times 1}(T'')))
    \label{equ7}
\end{equation}
\begin{equation}
    T^{(l)} = T''' + T^{(l-1)}
    \label{equ8}
\end{equation}

The $5\times5$ convolution kernels and dilations are the same as the $3\times3$ ones and thus are omitted.

\subsubsection{Global and Local Representation Alignment with Cross-granularity Contrastive Learning}

We use cross-granularity contrastive learning to align global sentence-level representations $h_{\texttt{cls}}\in R^d$, with local word-level representations $h_{\texttt{pos}}\in R^d (h_{\texttt{neg}}\in R^d)$. This method leverages positive and negative sample pairs to enhance the model's ability to distinguish between relevant and irrelevant representations. We construct two types of sample pairs:

\textbf{Positive sample pairs:} The global representation $h_{\texttt{cls}}$ is paired with the local representation $h_{\texttt{pos}}$ obtained by pooling the MMCNN output. The positive sample $h_{\texttt{pos}}$ can be defined as follows:
\begin{equation}
    h_{\texttt{pos}}=\frac{1}{n^2} \sum_{i=1}^{n} \sum_{j=1}^{n} T_{ij}
    \label{equ9}
\end{equation}
This represents the alignment between the global context and the local information within the same sentence.

\textbf{Negative sample pairs:} To construct the negative samples, we adopt the local representations $h_{\texttt{neg}}$ from different sentences in the same batch.

We define the cross-granularity contrastive learning loss using a margin-based ranking function. The goal is to minimize the distance between the global representation and the corresponding local representation (positive sample) while maximizing the distance between the global representation and the non-corresponding local representation (negative sample). The cross-granularity contrastive learning loss is defined as follows:
\begin{equation}
    L_{CL}=max(0,m+d(h_{\texttt{cls}}, h_{\texttt{neg}})-d(h_{\texttt{cls}}, h_{\texttt{pos}}))
    \label{equ10}
\end{equation}
where $d(\cdot)$ denotes to compute the Euclidean distance between two vectors, and $m$ (set to 1.0) is the margin that controls the separation between positive and negative sample pairs.

\subsubsection{Region Detection and Classification}

For each element $r^{(l)}_{ij}$ in the table representation $T^{(l)}$, classifiers calculate the probabilities of boundary tags being ‘S' and ‘E' using a sigmoid function:
\begin{equation}
    P^S_{ij} = \sigma(\text{Linear}(r^{(l)}_{ij}), \quad P^E_{ij} = \sigma(\text{Linear}r^{(l)}_{ij}),
    \label{equ11}
\end{equation}

Given a candidate region $S(a,b)$ and $E(c,d)$, the representation $r_{abcd}$ is constructed by concatenating the $S$, $E$ representations, and the max-pooling result of the region matrix:
\begin{equation}
  r_{abcd} = \left[ r^{(l)}_{ab}; r^{(l)}_{cd}; \text{MaxPool}\left( \{ r^{(l)}_{ij} \} \right) \right]
    \label{equ12}
\end{equation}

The sentiment polarity is predicted through a softmax function: $ SP \in \{\text{Positive}, \text{Negative}, \text{Neutral}, \text{Invalid}\} $.
\begin{equation}
     P_{abcd}(SP) = \text{Softmax}(\text{Linear}(r_{abcd}))
    \label{equ13}
\end{equation}
\subsubsection{Training and Decoding}
During training, the loss for boundary detection is calculated with binary cross-entropy:
\begin{equation}
     L_S = \text{BCEWithLogitsLoss}(P^S_{ij}, y^S_{ij})
    \label{equ14}
\end{equation}
\begin{equation}
     L_E = \text{BCEWithLogitsLoss}(P^E_{ij}, y^E_{ij})
      \label{equ15}
\end{equation}
where $y^S_{ij},  y^E_{ij}\in(0,1)$ are the ground truth boundary label.

For region classification, the ground truth region sentiment $SP^*$ is used to calculate loss with cross-entropy:
\begin{equation}
    L_{SP} = - \sum_{abcd} \log P_{abcd}(SP^*)
     \label{equ16}
\end{equation}

The total loss is:
\begin{equation}
    L_{Total} = L_{CL}+L_S+L_E+L_{SP}
     \label{equ17}
\end{equation}

During decoding, candidate regions are identified, and the sentiment polarity is predicted, yielding the triplet $(\text{aspect}, \text{opinion}, \text{sentiment})$.

\section{Experiments}
\subsection{Datasets}
This study assesses our model using four datasets presented by \cite{peng2020knowing}, namely 14Res, 14Lap, 15Res, and 16Res. These datasets encompass three from the restaurant domain and one from the laptop domain. The four datasets originate from the SemEval Challenges \cite{pontiki2016semeval} and were refined by Xu et al. \cite{xu2020position} based on the earlier version from Peng et al. \cite{peng2020knowing}. Table \ref{Table1} provides a summary of the detailed statistics for these benchmark datasets.

\begin{table}[htbp]
\centering
\caption{The dataset statistics are provided. `Sentence' refers to the total count of sentences, while `+', '0', and `-' represent the counts of positive, neutral, and negative sentiment triplets, respectively.}
\label{Table1}
\footnotesize
\begin{tabular}{|c|c|c|c|c|c|}
\hline
Datasets &   & Sentence & + & 0 & - \\ 
\hline 
\multirow{3}{*}{14Res}
& Train& 1266   & 1692 & 166 & 480 \\ 
\cline{2-6}
& Valid  & 310 & 404 & 54& 119 \\ 
\cline{2-6}
& Test & 492  & 773  & 66  & 155 \\ 
\hline   
 \multirow{3}{*}{14Lap}
& Train& 906 & 817 & 126 & 517 \\ 
\cline{2-6}
& Valid & 219 & 169   & 36 & 141 \\ 
\cline{2-6}
& Test  & 328  & 364  & 63 & 116 \\
\hline
\multirow{3}{*}{15Res}
& Train & 605  & 783   & 25 & 205 \\ 
\cline{2-6}
& Valid  & 148 & 185 & 11 & 53 \\ 
\cline{2-6}
& Test & 322  & 317  & 25 & 143 \\ 
\hline 
\multirow{3}{*}{16Res}
& Train & 857   & 1015  & 506 & 329 \\ 
\cline{2-6}
& Valid  & 210   & 252   & 11 & 76 \\ 
\cline{2-6}
& Test & 326  & 407  & 29 & 78 \\ 
\hline 
\end{tabular}
\end{table}

\begin{table*}[t]
\centering
\caption {Test set results for the ASTE task, with the highest values highlighted in bold. The first four results are from \cite{zhang2022boundary}, and the remaining data are sourced from the respective papers.}
\label{Table2}
\footnotesize
\begin{tabular}{|c|ccc|ccc|ccc|ccc|}
\hline
\multirow{2}{*}{Model} 
& \multicolumn{3}{c|}{14Res} & \multicolumn{3}{c|}{14Lap} & \multicolumn{3}{c|}{15Res} & \multicolumn{3}{c|}{16Res} \\ 
\cline{2-13} 
& \multicolumn{1}{c|}{$P.$} & $R.$ & $F1$ & \multicolumn{1}{c|}{$P.$} & $R.$ & $F1$ &  \multicolumn{1}{c|}{$P.$} & $R.$ & $F1$ &  \multicolumn{1}{c|}{$P.$} & $R.$ & $F1$  \\ 
\hline
GTS \cite{wu2020grid}  & 67.76 & 67.29 & 67.50 & 57.82 & 51.32 & 54.36 & 62.59 & 57.94 & 60.15 & 66.08 & 66.91 & 67.93 \\
\hline
Dual-Encoder \cite{jing2021seeking}  & 67.95 & 71.23 & 69.55 & 62.12 & 56.38 & 59.11 & 58.55 & 60.00 & 59.27 & 70.65 & 70.23 & 70.44 \\
\hline
EMC-GCN\cite{chen2022enhanced} & 71.21 & 72.39 & 71.78 & 61.70 & 56.26 & 58.81 & 61.54 & 62.47 & 61.93 & 65.62 & 71.30 & 68.33 \\
\hline
BDTF\cite{zhang2022boundary} & 75.53 & 73.24 & 74.35 & 68.94 & 55.97 & 61.74 & 68.76 & 63.71 & 66.12 & 71.44 & 73.13 & 72.27 \\
\hline
BMRC\cite{chen2021bidirectional} & 71.32 & 70.09 & 70.69 & 65.12 & 54.41 & 59.27 & 63.71 & 58.63 & 61.05 & 67.74 &68.56 & 68.13\\
\hline
BART-ASTE\cite{yan2021unified} & 65.52 & 64.99 & 65.25 & 61.41 & 56.19 & 58.69 & 59.14 & 59.38 & 59.26 & 66.6 &68.68 & 67.62\\
\hline
Span ASTE (POS\&CL)\cite{li2024improving}& 73.33 & \textbf{76.31} & 74.79 & 65.20 & \textbf{60.18} & 62.59 & 68.39 & 61.97 & 65.03 & 71.02 &71.31 & 71.17 \\
\hline
BTF-CCL (Ours) & \textbf{80.41}& 71.83 & \textbf{75.88} & \textbf{70.66} & 57.31 & \textbf{63.29} & \textbf{71.66} & \textbf{64.12} & \textbf{67.68} & \textbf{72.56} & \textbf{75.10} & \textbf{73.80} \\
\hline
\end{tabular}
\end{table*}
\subsection{Implementation Details}
We employ the pre-trained ``BERT-base-uncased" model, which consists of 110 million parameters, 12 attention heads, 12 hidden layers, and a hidden size of 768, as our encoding layer. The model is trained for 10 epochs and the best model parameters are selected based on the highest F1 score on the validation set, which are then applied to evaluate the model’s performance on the test set.

\subsection{Baselines}

\textbf{Table-filling methods} frame aspect terms, opinion terms, and their sentiment as word-pair interactions. \cite{wu2020grid} presented the Grid Tagging Scheme (GTS), coupled with an inference technique to exploit interdependencies between opinion elements. \cite{jing2021seeking} proposed the Dual-Encoder to leverage table-sequence encoders for capturing both sequence and table representations. \cite{chen2022enhanced} developed the Enhanced Multi-Channel Graph Convolutional Network (EMC-GCN) to incorporate linguistic features. \cite{zhang2022boundary} introduced the boundary-driven table-filling method (BDTF), emphasizing word relationships.

\textbf{Other end-to-end methods} handle triplet extraction by learning a joint model. \cite{chen2021bidirectional} proposed a bidirectional machine reading comprehension (BMRC) framework that uses three query types to capture links between subtasks. \cite{li2024improving} presented a span-level ASTE method with a POS filter and contrastive learning, further improving model performance. \cite{yan2021unified} reformulated all ABSA subtasks into a unified generative approach utilizing BART.

\subsection{Main Results}

The results in Table \ref{Table2} show that our model outperforms the baseline models in terms of Precision (P), Recall (R), and F1 scores across the four datasets. Specifically, it improves the F1 score by 1.09 and 0.7 points on the 14Res and 14Lap datasets, and by 1.56 and 1.53 points on the 15Res and 16Res datasets, respectively, compared to the previous best result from Span ASTE (POS\&CL) \cite{li2024improving} and BDTF \cite{zhang2022boundary}. While Span ASTE (POS\&CL) demonstrate higher recall on the 14Res and 14Lap datasets compared to our model, it is worth highlighting that our model excels in precision. This suggests that our BTF-CCL maintains a better balance between precision and recall, resulting in enhanced overall performance.

\subsection{Ablation Study}
We performed an ablation study to evaluate our model's performance under various configurations. The results, shown in Table \ref{Table3}, summarize the outcomes for each setup. The details of the ablation experiments and their analysis are as follows:

\textbf{Contrastive learning:} Without contrastive learning, the semantic consistency between the global representation and the local representation is lost. The findings from this ablation experiment indicate a significant decline in the model's performance across all datasets. This suggests that contrastive learning enhances the guidance of sentence-level global context over word-level local features.

\textbf{MMCNN:} The MMCNN is used to capture rich semantic information across different levels of granularity. To verify its effectiveness, we removed it in the ablation experiment, and the results show that the F1 scores of the model decreased across all four datasets without MMCNN.

\begin{table}[htbp]
\centering
\caption{The results of ablation experiment. w/o denotes removing the component.}
\label{Table3}
\begin{tabular}{|c|c|c|c|c|}
\hline
Model & 14Res & 14Lap & 15Res & 16Res \\ 
\hline
BTF-CCL & \textbf{75.88} & \textbf{63.29} & \textbf{67.68} & \textbf{73.80}  \\ 
\hline
w/o CCL & 74.57 & 62.01 & 66.88 & 73.05  \\ 
\hline
w/o MMCNN & 75.01 &62.36  &  67.17& 73.35 \\ 
\hline
\end{tabular}
\end{table}

\section{Conclusion}
In this work, we propose boundary-driven table-filling with cross-granularity contrastive learning (BTF-CCL) to enhance the semantic consistency between sentence-level representations and word-level representations. By constructing positive and negative sample pairs, the model is forced to learn the associations at both the sentence level and the word level. Additionally, a multi-scale, multi-granularity convolutional method is proposed to capture rich semantic information better. Our approach can capture sentence-level contextual information
more effectively while maintaining sensitivity to local details. These contributions together enhanced the performance of table-filling ASTE models, highlighting the effectiveness of our approach in tackling key challenges and advancing sentiment analysis in aspect-based tasks.


\vspace{12pt}
\bibliographystyle{IEEEtran}
\bibliography{reference}

\begin{thebibliography}{10}
\providecommand{\url}[1]{#1}
\csname url@samestyle\endcsname
\providecommand{\newblock}{\relax}
\providecommand{\bibinfo}[2]{#2}
\providecommand{\BIBentrySTDinterwordspacing}{\spaceskip=0pt\relax}
\providecommand{\BIBentryALTinterwordstretchfactor}{4}
\providecommand{\BIBentryALTinterwordspacing}{\spaceskip=\fontdimen2\font plus
\BIBentryALTinterwordstretchfactor\fontdimen3\font minus
  \fontdimen4\font\relax}
\providecommand{\BIBforeignlanguage}[2]{{%
\expandafter\ifx\csname l@#1\endcsname\relax
\typeout{** WARNING: IEEEtran.bst: No hyphenation pattern has been}%
\typeout{** loaded for the language `#1'. Using the pattern for}%
\typeout{** the default language instead.}%
\else
\language=\csname l@#1\endcsname
\fi
#2}}
\providecommand{\BIBdecl}{\relax}
\BIBdecl

\bibitem{pontiki-etal-2014-semeval}
\BIBentryALTinterwordspacing
M.~Pontiki, D.~Galanis, J.~Pavlopoulos, H.~Papageorgiou, I.~Androutsopoulos,
  and S.~Manandhar, ``{S}em{E}val-2014 task 4: Aspect based sentiment
  analysis,'' in \emph{Proceedings of the 8th International Workshop on
  Semantic Evaluation ({S}em{E}val 2014)}, P.~Nakov and T.~Zesch, Eds.\hskip
  1em plus 0.5em minus 0.4em\relax Dublin, Ireland: Association for
  Computational Linguistics, Aug. 2014, pp. 27--35. [Online]. Available:
  \url{https://aclanthology.org/S14-2004}
\BIBentrySTDinterwordspacing

\bibitem{pontiki-etal-2015-semeval}
\BIBentryALTinterwordspacing
M.~Pontiki, D.~Galanis, H.~Papageorgiou, S.~Manandhar, and I.~Androutsopoulos,
  ``{S}em{E}val-2015 task 12: Aspect based sentiment analysis,'' in
  \emph{Proceedings of the 9th International Workshop on Semantic Evaluation
  ({S}em{E}val 2015)}, P.~Nakov, T.~Zesch, D.~Cer, and D.~Jurgens, Eds.\hskip
  1em plus 0.5em minus 0.4em\relax Denver, Colorado: Association for
  Computational Linguistics, Jun. 2015, pp. 486--495. [Online]. Available:
  \url{https://aclanthology.org/S15-2082}
\BIBentrySTDinterwordspacing

\bibitem{pontiki2016semeval}
M.~Pontiki, D.~Galanis, H.~Papageorgiou, I.~Androutsopoulos, S.~Manandhar,
  M.~Al-Smadi, M.~Al-Ayyoub, Y.~Zhao, B.~Qin, O.~De~Clercq \emph{et~al.},
  ``Semeval-2016 task 5: Aspect based sentiment analysis,'' in
  \emph{International workshop on semantic evaluation}, 2016, pp. 19--30.

\bibitem{manek2017aspect}
A.~S. Manek, P.~D. Shenoy, and M.~C. Mohan, ``Aspect term extraction for
  sentiment analysis in large movie reviews using gini index feature selection
  method and svm classifier,'' \emph{World wide web}, vol.~20, pp. 135--154,
  2017.

\bibitem{li2018aspect}
X.~Li, L.~Bing, P.~Li, W.~Lam, and Z.~Yang, ``Aspect term extraction with
  history attention and selective transformation,'' \emph{arXiv preprint
  arXiv:1805.00760}, 2018.

\bibitem{ma2019exploring}
D.~Ma, S.~Li, F.~Wu, X.~Xie, and H.~Wang, ``Exploring sequence-to-sequence
  learning in aspect term extraction,'' in \emph{Proceedings of the 57th annual
  meeting of the association for computational linguistics}, 2019, pp.
  3538--3547.

\bibitem{akhtar2020multi}
M.~S. Akhtar, T.~Garg, and A.~Ekbal, ``Multi-task learning for aspect term
  extraction and aspect sentiment classification,'' \emph{Neurocomputing}, vol.
  398, pp. 247--256, 2020.

\bibitem{wang2019transferable}
W.~Wang and S.~J. Pan, ``Transferable interactive memory network for domain
  adaptation in fine-grained opinion extraction,'' in \emph{Proceedings of the
  aaai conference on artificial intelligence}, vol.~33, no.~01, 2019, pp.
  7192--7199.

\bibitem{wu2020deep}
M.~Wu, W.~Wang, and S.~J. Pan, ``Deep weighted maxsat for aspect-based opinion
  extraction,'' in \emph{Proceedings of the 2020 conference on empirical
  methods in natural language processing (EMNLP)}, 2020, pp. 5618--5628.

\bibitem{dai2022reasoning}
Y.~Dai, P.~Wang, and X.~Zhu, ``Reasoning over multiplex heterogeneous graph for
  target-oriented opinion words extraction,'' \emph{Knowledge-Based Systems},
  vol. 236, p. 107723, 2022.

\bibitem{feng2021target}
Y.~Feng, Y.~Rao, Y.~Tang, N.~Wang, and H.~Liu, ``Target-specified sequence
  labeling with multi-head self-attention for target-oriented opinion words
  extraction,'' in \emph{Proceedings of the 2021 conference of the north
  american chapter of the association for computational linguistics: Human
  language technologies}, 2021, pp. 1805--1815.

\bibitem{peng2020knowing}
H.~Peng, L.~Xu, L.~Bing, F.~Huang, W.~Lu, and L.~Si, ``Knowing what, how and
  why: A near complete solution for aspect-based sentiment analysis,'' in
  \emph{Proceedings of the AAAI conference on artificial intelligence},
  vol.~34, no.~05, 2020, pp. 8600--8607.

\bibitem{xu2020position}
L.~Xu, H.~Li, W.~Lu, and L.~Bing, ``Position-aware tagging for aspect sentiment
  triplet extraction,'' \emph{arXiv preprint arXiv:2010.02609}, 2020.

\bibitem{zhang2020multi}
C.~Zhang, Q.~Li, D.~Song, and B.~Wang, ``A multi-task learning framework for
  opinion triplet extraction,'' \emph{arXiv preprint arXiv:2010.01512}, 2020.

\bibitem{xu2021learning}
L.~Xu, Y.~K. Chia, and L.~Bing, ``Learning span-level interactions for aspect
  sentiment triplet extraction,'' \emph{arXiv preprint arXiv:2107.12214}, 2021.

\bibitem{wu2020grid}
Z.~Wu, C.~Ying, F.~Zhao, Z.~Fan, X.~Dai, and R.~Xia, ``Grid tagging scheme for
  aspect-oriented fine-grained opinion extraction,'' \emph{arXiv preprint
  arXiv:2010.04640}, 2020.

\bibitem{li2024improving}
Q.~Li, W.~Wen, and J.~Qin, ``Improving span-based aspect sentiment triplet
  extraction with part-of-speech filtering and contrastive learning,''
  \emph{Neural Networks}, vol. 177, p. 106381, 2024.

\bibitem{jing2021seeking}
H.~Jing, Z.~Li, H.~Zhao, and S.~Jiang, ``Seeking common but distinguishing
  difference, a joint aspect-based sentiment analysis model,'' \emph{arXiv
  preprint arXiv:2111.09634}, 2021.

\bibitem{zhang2022boundary}
Y.~Zhang, Y.~Yang, Y.~Li, B.~Liang, S.~Chen, Y.~Dang, M.~Yang, and R.~Xu,
  ``Boundary-driven table-filling for aspect sentiment triplet extraction,'' in
  \emph{Proceedings of the 2022 Conference on Empirical Methods in Natural
  Language Processing}, 2022, pp. 6485--6498.

\bibitem{sun2024rethinking}
Q.~Sun, L.~Yang, M.~Ma, N.~Ye, and Q.~Gu, ``Rethinking aste: A minimalist
  tagging scheme alongside contrastive learning,'' \emph{arXiv preprint
  arXiv:2403.07342}, 2024.

\bibitem{devlin2018bert}
J.~Devlin, ``Bert: Pre-training of deep bidirectional transformers for language
  understanding,'' \emph{arXiv preprint arXiv:1810.04805}, 2018.

\bibitem{chen2022enhanced}
H.~Chen, Z.~Zhai, F.~Feng, R.~Li, and X.~Wang, ``Enhanced multi-channel graph
  convolutional network for aspect sentiment triplet extraction,'' in
  \emph{Proceedings of the 60th Annual Meeting of the Association for
  Computational Linguistics (Volume 1: Long Papers)}, 2022, pp. 2974--2985.

\bibitem{chen2021bidirectional}
S.~Chen, Y.~Wang, J.~Liu, and Y.~Wang, ``Bidirectional machine reading
  comprehension for aspect sentiment triplet extraction,'' in \emph{Proceedings
  of the AAAI conference on artificial intelligence}, vol.~35, no.~14, 2021,
  pp. 12\,666--12\,674.

\bibitem{yan2021unified}
H.~Yan, J.~Dai, X.~Qiu, Z.~Zhang \emph{et~al.}, ``A unified generative
  framework for aspect-based sentiment analysis,'' \emph{arXiv preprint
  arXiv:2106.04300}, 2021.

\end{thebibliography}

\end{document}